%% file: main.tex
\documentclass[10pt,twocolumn,letterpaper]{article}

\usepackage{wacv}      % To produce the REVIEW version for the algorithms track
\usepackage{graphicx}
\usepackage{amsmath}
\usepackage{amssymb}
\usepackage{booktabs}
\usepackage{pifont}
\usepackage{todonotes}
\usepackage[pagebackref,breaklinks,colorlinks]{hyperref}
\usepackage{tabularx}
\usepackage{hhline}
\usepackage[capitalize]{cleveref}
\crefname{section}{Sec.}{Secs.}
\Crefname{section}{Section}{Sections}
\Crefname{table}{Table}{Tables}
\crefname{table}{Tab.}{Tabs.}

\def\wacvPaperID{1452} % *** Enter the WACV Paper ID here
\def\confName{WACV}
\def\confYear{2024}

\begin{document}

%\title{Low Cost Construction of Indoor-Outdoor Short-Long Range Skeleton Gait Dataset with Hybrid 3D Computer Vision and Learning}
\title{DIOR: Dataset for Indoor-Outdoor Reidentification - Long Range 3D/2D Skeleton Gait Collection Pipeline, Semi-Automated Gait Keypoint Labeling and Baseline Evaluation Methods}

\author{Yuyang Chen, Praveen Raj Masilamani, Bhavin Jawade, Srirangaraj Setlur, Karthik Dantu\\
Department of Computer Science and Engineering\\
University at Buffalo, SUNY\\
{\tt\small \{yuyangch, pmasilam, bhavinja, setlur, kdantu\}@buffalo.edu}
}
\maketitle

%%%%%%%%% ABSTRACT
\begin{abstract}
In recent times, there is an increased interest in the identification and re-identification of people at long distances, such as from rooftop cameras, UAV cameras, street cams, and others. Such recognition needs to go beyond face and use whole-body markers such as gait. However, datasets to train and test such recognition algorithms are not widely prevalent, and fewer are labeled. This paper introduces DIOR - a framework for data collection, semi-automated annotation, and also provides a dataset with 14 subjects and 1.649 million RGB frames with 3D/2D skeleton gait labels, including 200 thousands frames from a long range camera. Our approach leverages advanced 3D computer vision techniques to attain pixel-level accuracy in indoor settings with motion capture systems. Additionally, for outdoor long-range settings, we remove the dependency on motion capture systems and adopt a low-cost, hybrid 3D computer vision and learning pipeline with only 4 low-cost RGB cameras, successfully achieving precise skeleton labeling on far-away subjects, even when their height is limited to a mere 20-25 pixels within an RGB frame. On publication, we will make our pipeline open for others to use. 
\end{abstract}

%%%%%%%%% BODY TEXT
\input{intro}
\input{relwork}
\input{approach}

\input{eval}

\input{dataset}
\input{conc}

%------------------------------------------------------------------------

% You must include your signed IEEE copyright release form when you submit your finished paper.
% We MUST have this form before your paper can be published in the proceedings.

% Please direct any questions to the production editor in charge of these proceedings at the IEEE Computer Society Press:
% \url{https://www.computer.org/about/contact}.

%%%%%%%%% REFERENCES
{\small
\bibliographystyle{ieee_fullname}
\bibliography{egbib}
}

\end{document}

%% file: intro.tex
\begin{table*}[h!!]
  \centering
  \begin{tabular}{|c|c|c|c|c|c|c|}
    \hline
    Dataset &MoCap Pose Gallery &Multi-View & 2D-pose& 3D-pose & long range& $<$25 pixel height  \\
\specialrule{1.5pt}{0pt}{0pt} 
    BRIAR~\cite{cornett2023expanding}& \color{red}\ding{55}& \color{green}\ding{51}& \color{red}\ding{55} & \color{red}\ding{55} & \color{green}\ding{51} &\color{red}\ding{55}\\
    Dronesurf~\cite{kalra2019dronesurf}& \color{red}\ding{55} & \color{red}\ding{55}& \color{red}\ding{55}& \color{red}\ding{55} & \color{green}\ding{51} &\color{red}\ding{55}\\
    OUMVLP-pose~\cite{takemura2018multi}& \color{red}\ding{55}& \color{green}\ding{51}& \color{green}\ding{51} & \color{red}\ding{55} & \color{red}\ding{55} &\color{red}\ding{55}\\
    CASIA-B~\cite{yu2006framework}& \color{red}\ding{55}& \color{green}\ding{51}& \color{green}\ding{51} & \color{red}\ding{55} & \color{red}\ding{55} &\color{red}\ding{55}\\
    Gait3D~\cite{zheng2022gait}& \color{red}\ding{55}& \color{red}\ding{55}& \color{green}\ding{51}~ & \color{green}\ding{51} & \color{red}\ding{55} &\color{red}\ding{55}\\
    \hline
    \textbf{DIOR}(ours)& \color{green}\ding{51}\color{black}* & \color{green}\ding{51}& \color{green}\ding{51}& \color{green}\ding{51} & \color{green}\ding{51}& \color{green}\ding{51}\\

    \hline
    
  \end{tabular}
  \caption{DIOR has two unique features. First, the inclusion of long range, 20-25 pixels, extremely low resolution subject data. Second, the utilization of a Motion Capture system for pixel-accurate 3D/2D pose for half the gallery. * Around half, or 800k frames of DIOR pose data are derived with the help of an indoor VICON system. The other 800k RGB frames are captured outdoor, with 4 Realsense D455 cameras.}
  \vspace{-15pt}
  \label{tab:1}
  
\end{table*}
\section{Introduction}
\label{sec:intro}

\begin{figure}[h]
    \centering
    \includegraphics[width=0.9\linewidth]{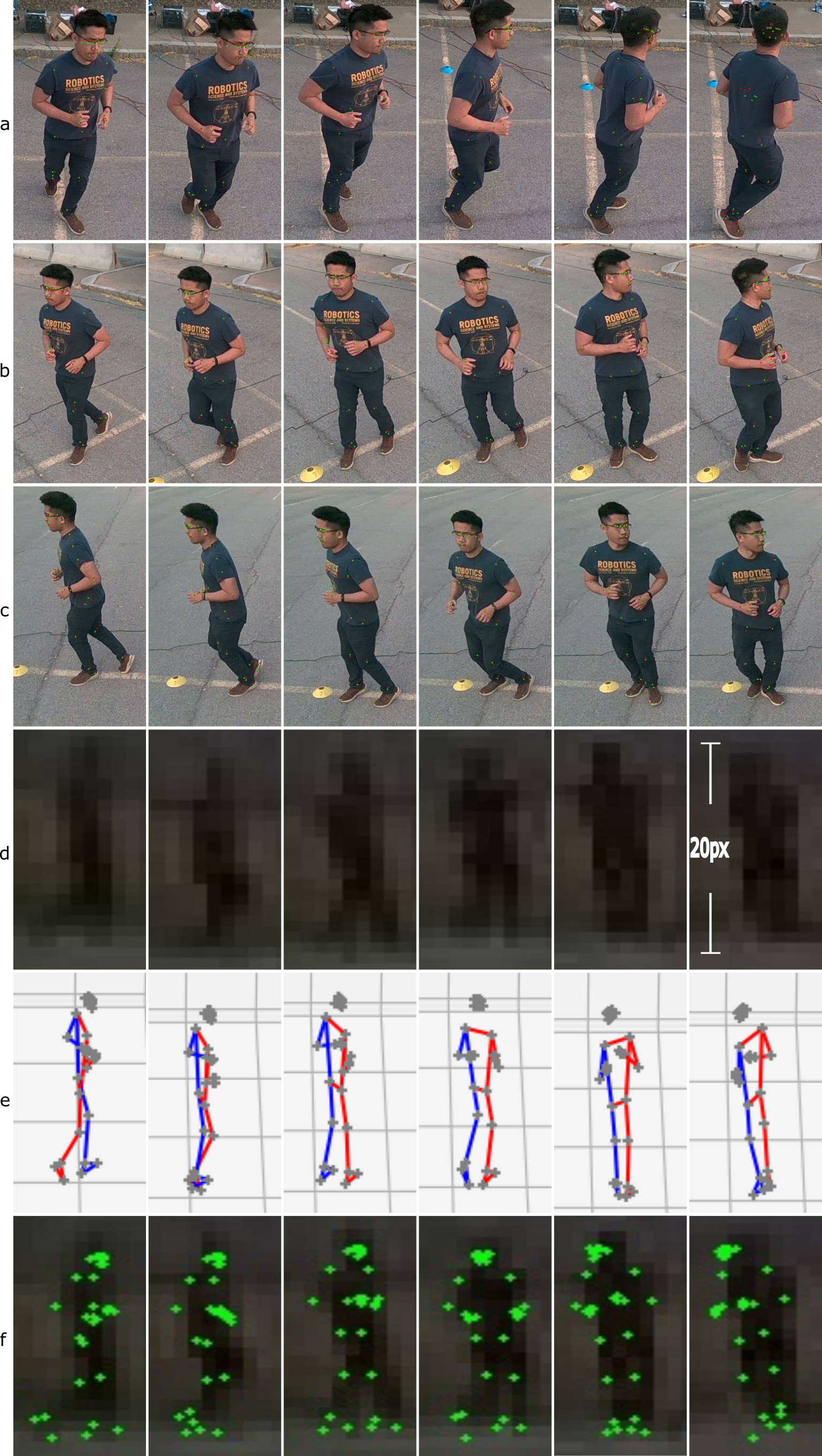}
    \caption{Cam view row d) is placed exactly opposite to c), hence images appear to be horizontally flipped. see Figures ~\ref{fig:system} and~\ref{fig:long_range_cam_adjustment}. a),b),c)-Closed range cam with reprojected 2d pose labels in green dots (zoom in). d)Long range view e) reprojected gait keypoints.} %f) Long range camera 2 RGB image with reprojected 2d pose labels.} 
    \vspace{-20pt} \label{fig:first_page_first_image}
\end{figure}

Anthropomorphic features such as gait are of increased interest for various applications such as activity recognition, identity recognition and others. An increasingly interesting aspect of this problem is gait recognition in both indoor and outdoor settings. Lighting, viewing angle, proximity and several other aspects are vastly different when seeing a person indoors vs outdoors. A large body of research looks at indoor, close-range images for gait and identity recognition but not from long range. Further, it is more challenging to have general pipelines that perform such tasks for both short and long range. 

This paper contributes to the advancement of skeleton walking gait detection and recognition. Our primary contribution is the development of a comprehensive dataset, crucial for training and evaluating robust algorithms for gait-based tasks both indoors and outdoors. A second challenge in such datasets is accurate annotation of gait keypoints. To this end, we employ a semi-automatic annotation process that enables efficient and precise annotations at high speed (sub-second per frame).

In indoor settings, we leverage a Motion Capture system (MoCap) and 3D computer vision techniques to achieve pixel-level accurate labeling of 2D Gait Key-points. By leveraging accurate 3D gait keypoint information from motion capture systems and perspective geometry, we are able to accurately localize our RGBD camera array, and re-project precise 2d gait keypoints onto RGB frames.  The accuracy and reliability demonstrated by our method in indoor scenarios have promising implications for applications such as {\it 2D/3D pose detection and multi-view 2D/3D gait recognition}. 

\begin{figure*}[h]
    \centering\includegraphics[width=0.75\textwidth]{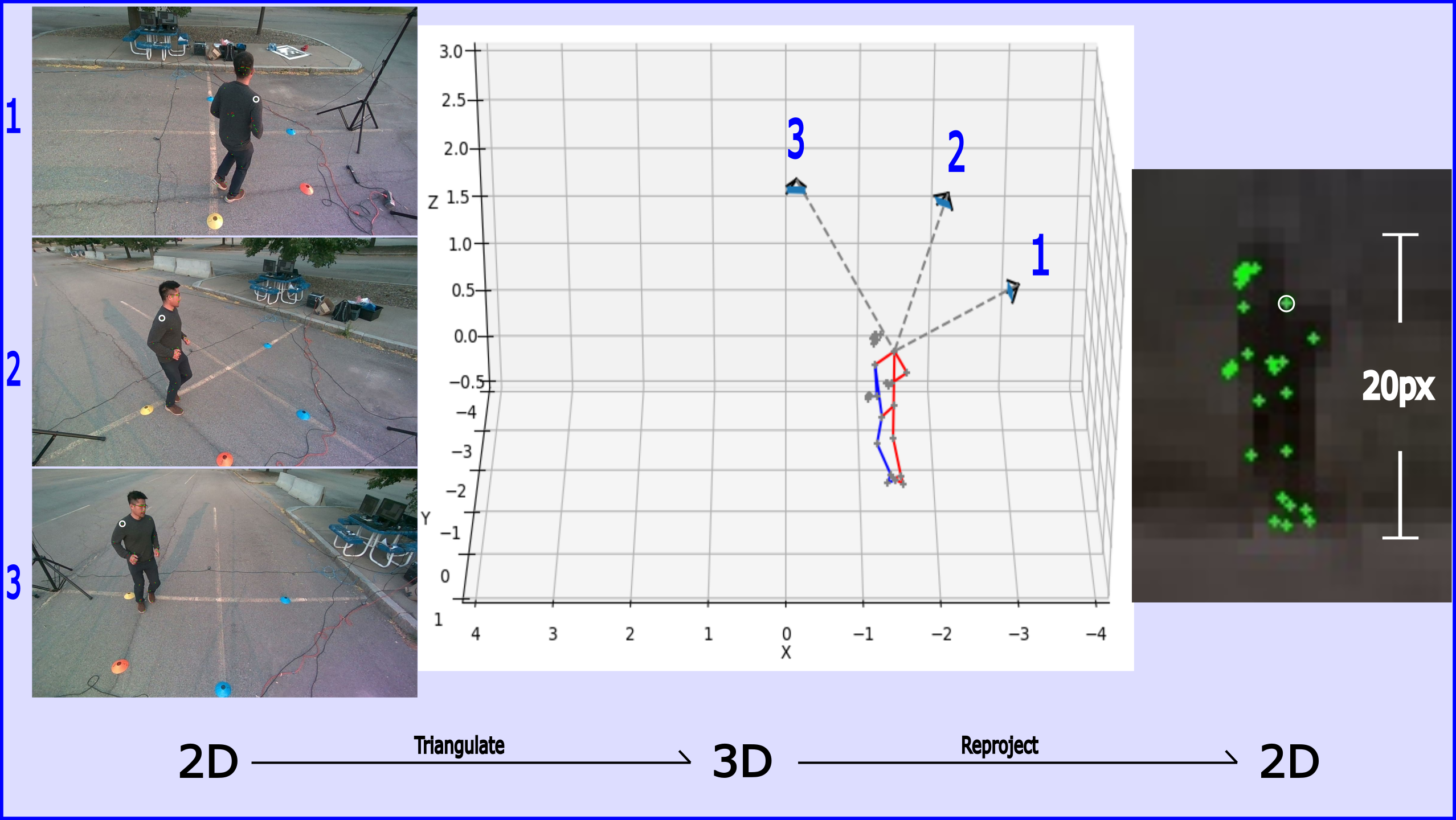}
    \caption{(1) MLKIT marks the 2D keypoints on all 3 RGB images, (2) we triangulate each to obtain 3D gait key-points, and (3) reproject the 3D gait onto the long range view at 60 meters.}
    \label{fig:system}
    \vspace{-15pt}
\end{figure*}

However, outdoor settings present additional challenges. Setting up motion capture systems in the outdoor settings is generally cost prohibitive. We do away with relying on the motion capture system, and use 4 RGB cameras for similar annotation. Further  more, varying lighting conditions, occlusions, and the distant nature of the subjects add considerable difficulties. To address these complexities, we have developed a hybrid pipeline that combines the strengths of 3D Computer Vision and learning methodologies. We place 3 RGB cameras in the close range, for existing learning methods to identify 2d gait keypoints on RGB images. We place 1 RGB camera at long range. With multi-view, 2D skeleton keypoint labels from existing learning methods, we triangulate 3D skeleton keypoints. We then re-project the 3D skeletons onto the long range camera frames. We achieve skeleton labeling on long range subjects, {\bf even when the subject occupies a limited 20-25 pixel within an RGB frame}. Further, utilizing the 3D Computer Vision, we can create dataset with partially occluded subjects in the long range frames.  Sample images are shown in \autoref{fig:first_page_first_image}. Camera view row d) is placed exactly opposite to c), hence images appear to be horizontally flipped. see Figure~\ref{fig:system} and~\ref{fig:long_range_cam_adjustment} for placement details.

This progress brings us closer to realizing long-range gait detection in outdoor environments, potentially benefiting {\it perimeter security, public safety, and autonomous driving applications}. %citations

%\color{red}
%Chen:Consider deleting this paragraph to save some space? almost over 8 pages.
%The research and accompanying open-source code presented in this paper contributes to the scientific understanding and progress of skeleton walking gait detection. By addressing the challenges posed by indoor and outdoor scenarios, we seek to establish new benchmarks and facilitate the development of more robust and generalizable algorithms. It is our hope that this research will inspire further exploration and innovation in skeleton walking gait detection, fostering the advancement of human recognition systems and enabling practical applications across various domains.
%\color{black}

%The subsequent sections of this paper provide detailed explanations of our methodologies. In Section~~\ref{sec:methods} we present our methods for semi-automatic annotations, and the utilization of 3D computer vision and learning techniques in both indoor and outdoor settings.  In Section~\ref{sec:eval}, we present quantitative metrics on the quality of our 2D/3D labels. In Section~\ref{sec:baseline} we present baseline results of existing RGB-to-Pose and Gait Recognition pipelines. In Section~\ref{sec:dataset}, we present detail statistics and protocol of our dataset.  
%\todo[inline]{check if these are correct. Ideally, you use section refs rather than numbers like this. Chen: added section refs}

Our work makes the following contributions: 
%\begin{itemize}
%    \item \textbf{Dataset:} dual use -- person re-identification and RGB to 2d gait dataset. For its details and statistics see Table~\ref{tab:1} and Section~\ref{sec:dataset}.
%    \item \textbf{Semi Automated annotation methods:} A novel 3D Computer Vision method for indoor MoCap Assisted environment to reproject and auto-annotate MoCap 3d gait onto RGB camera frames. And a Novel hybrid 3D Computer Vison and Learning method that can collect 3D/2D pose in the outdoor environment with 4 low-cost RGB cameras. 
%    \item \textbf{Open Source code:} A github repository of our collection and semi-auto labeling code upon publication.
%\end{itemize}
\begin{itemize}
    \item \textbf{Dataset:} A novel dataset that captures subjects indoors and outdoors, with two different sets of clothing each. This includes a gallery with images from multiple angles and heights. The outdoor dataset also has long range images where the subject is less than 25 pixels in size. Details are described in \autoref{tab:1}. 
    \item \textbf{Semi Automated annotation pipeline:} A novel pipeline for indoor MoCap assisted environment to reproject and auto-annotate MoCap 3d gait onto RGB camera frames. And a novel vision pipeline that can annotate 3D/2D gait keypoints in the outdoor environment with 4 low-cost RGB cameras. 
    \item \textbf{Baseline Evaluation:} As a baseline, we test our dataset with multiple gait-based recognition papers for demonstration. This can help set a baseline for future advancements in this area using the DIOR dataset. 
\end{itemize}

{\bf All the data collection performed for this dataset are in compliance with our approved IRB. }

%----------------------------------------, , , ,.---------------------------------

%% file: relwork.tex
\section{Related Work}
\label{sec:related_work}
\noindent \textbf{Gait Recognition Dataset:}
There is a large collection of gait recognition data-sets available including and not limited to BRIAR~\cite{cornett2023expanding}, Gait3D~\cite{zheng2022gait}, Dronesurf~\cite{kalra2019dronesurf}, OUMVLP~\cite{takemura2018multi}, CASIA-B~\cite{yu2006framework}. To the best of our knowledge, none of the existing datasets contain long range skeleton labels where subjects appear less than 25 pixels in height.  The Dior dataset has long range image with low pixel count subject and 3D/2D skeleton labels. A detailed comparison of features is in Table ~\ref{tab:1}. DIOR has two unique features. First, the inclusion of long range, 20-25 pixel, extremely low resolution data. Second, the utilization of Motion Capture system for pixel-accuracy indoor settings for half the gallery.

\noindent \textbf{RGB to 2D pose:}
AlphaPose~\cite{fang2022alphapose}, OpenPose~\cite{cao2017realtime} and YOLOv7~\cite{wang2022yolov7} are trained on COCO ~\cite{lin2014microsoft}. COCO contains 66,000+ images of various environment and walking posture with annotated 2d pose. In comparison, our dataset has 800,000+ images with motion capture 2d pose annotations, but limited to indoor environment. 

\noindent \textbf{Automatic 3D/2D pose Annotation and long range data:}~Typically, one or more learning based pipelines are used for automatic 2D pose annotation, namely AlphaPose~\cite{fang2022alphapose} and OpenPose~\cite{cao2017realtime}. Moreover recent work like YOLOv7~\cite{wang2022yolov7} can also help on these tasks. The use of such models are limited to certain range and minimum subject resolutions, and as the subjects appear further away, and span fewer pixels, they are no longer effective. We will demonstrate this point through our RGB-to-2D-pose baseline in~\ref{subsec:rgb_to_pose}. This results in an interesting phenomenon in existing datasets -- The datasets with long range data do not have gait pose labels, and the datasets with gait pose labels do not have long range data. DIOR however, has both long range data, and 3D/2D pose labels. 

\noindent \textbf{3D Computer Vision:} 
Uniquely, DIOR leverages knowledge from multiple-views geometry~\cite{hartley2003multiple}, in accurately localizing cameras for both their orientation and translation.  We use OpenCV~\cite{bradski2000opencv}'s PnP~\cite{gao2003complete}~\cite{lepetit2009ep} library, in motion capture settings; We use GTSAM~\cite{gtsam}~\cite{factor_graphs_for_robot_perception}'s bundle adjustment (SfM) code base in the outdoor settings with only RGB cameras. Having the accurate camera extrinsics, we can then triangulate and re-project the 3D and 2D points. We will go over these details in section~\ref{sec:methods}. 

\noindent \textbf{Skeleton Based Person ReID:}
In the proposed work we evaluate skeleton (keypoint) based gait recognition methods from recent works. We evaluate a homogenous multi-axial mixer called GaitFormer ~\cite{pinyoanuntapong2023gaitmixer}. Along with this we also evaluate GaitMixer~\cite{pinyoanuntapong2023gaitmixer} which is a hetrogenous multi-directional mixer design, combining a spatial self-attention mixer and a temporal large-kernel convolution mixer. This combination enables the model to capture intricate multi-frequency patterns within gait feature maps. . ~\cite{pinyoanuntapong2023gaitmixer} showed that though Gaitformer cannot model high frequency components, GaitMixer can concentrate on both high-frequency and low-frequency components along both temporal and spatial axes in feature maps. In contrast to attention based methods, we also evaluate graph based architectures on DIOR. GaitGraph~\cite{teepe2021gaitgraph} combines skeleton poses with Graph Convolutional Network (GCN) to more discriminative person features. 
\vspace{-0.7em}
\color{black}
% \subsection{Indoor Short Range}
% To train the above models, accurately label datasets are needed. A large collections of datasets provides sillouettte only labels \color{blue}cite 5 works here\color{black}. And another set of datasets provides skeleton labels \color{blue}cite 3 works here\color{black}. Notably, the skeleton datasets are either manually labeled, which is costly, or labeled with OpenPose. 
% \subsubsection{MLKIT Learning Pipeline}
% Empirically, in our experiment, MLKIT is the most reliable pipeline in detecting short range subjects. We leverage MLKIT for as part of our outdoor long range pipeline

% \subsection{Outdoor Long Range}
% There is a lack of long range dataset with any gait labels, sillouette or skeleton. 
% Further, none of the existing models, OpenPose or MLKIT included, perform well in the long range settings. We will quantitatively demonstrate this point in Section~\ref{sec:results}. Therefore, to the best of our knowledge, automatic labelling with existing pipeline is not possible. 

% Further more, critically, due to the long range nature of the imagery, subjects appears in very limited amount of pixels(\color{blue}show an image here\color{black}), renders manual label challenging and costly.  

%% file: approach.tex
\section{Approach}
\label{sec:methods}
Our work is in two parts - semi-automated annotation for indoor and outdoor data. The high level ideas are similar (shown in \autoref{fig:system}) -- establishing a camera array where camera extrinsic parameters (position and orientation) are accurately estimated, then triangulate 3D points based on the 2D points on a portion of the cameras. Then reproject the 3D points to the other portion of the camera's frames as 2D labels. \\
NOTE: Estimation of the target group of RGB camera extrinsics and associated corrections form the bulk of the manual part of the pipeline. 

\subsection{Indoor Setup and Annotation Pipeline}

\begin{figure}[h]
    \centering
    \includegraphics[width=0.9\linewidth]{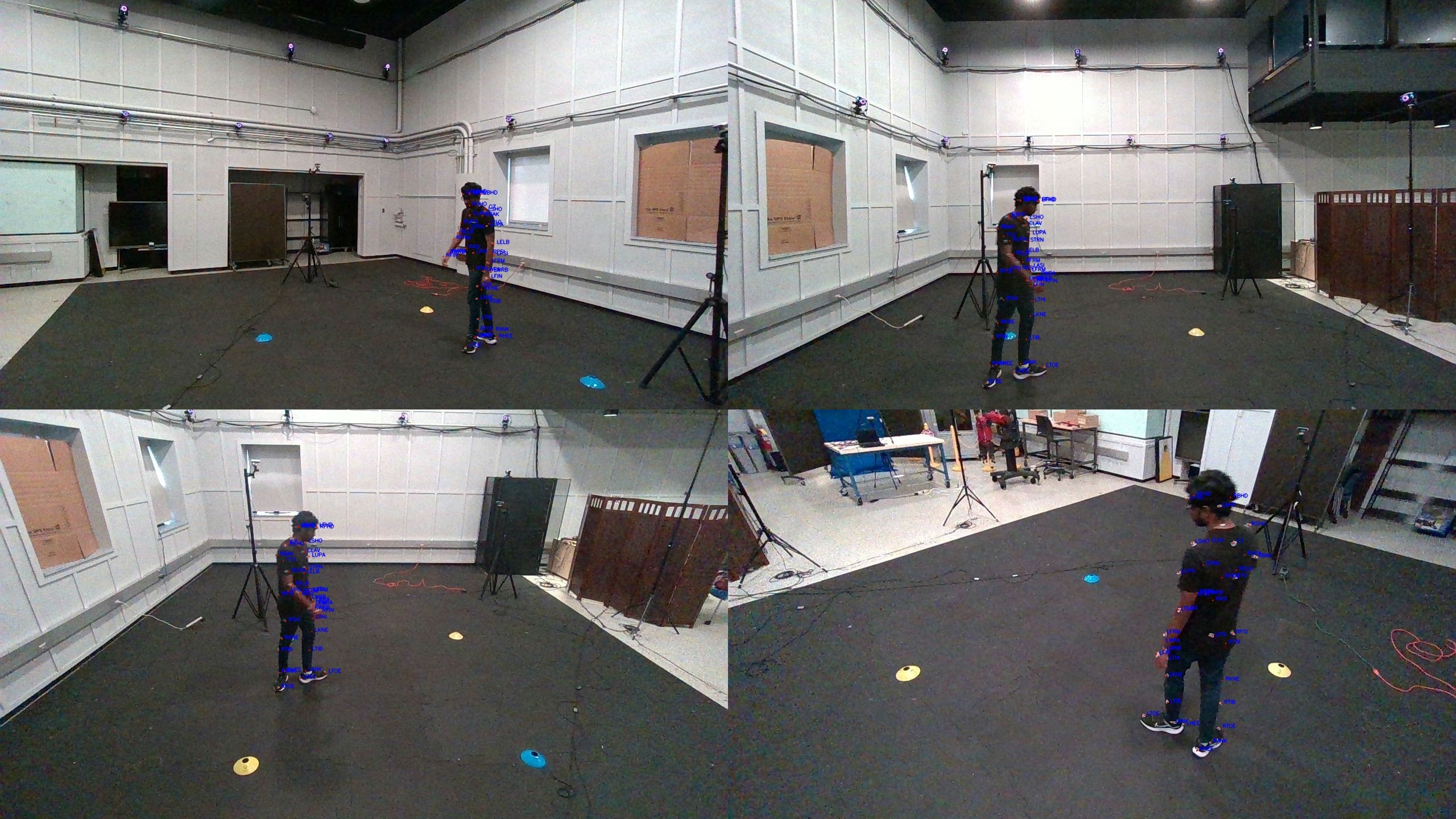}
    \caption{Indoor RGB images from 4 cameras, with 2d gait keypoints labeled (in blue). Zoom in to view each label.}
    \vspace{-7pt}\label{fig:4_cam_view_indoor}
\end{figure}
A motion capture system consists of multiple IR cameras and can accurately estimate the position and orientation of its own camera array, then triangulate any IR reflective markers in its capture volume (mm accuracy). The subjects wear 33 markers according to {\it Vicon Plug-in gait} spec. The system can capture the markers with labels at 100Hz, which results in mm level accurate 3D pose data. Therefore, the system already covers the 2D-3D part of the 2D-3D-2D work flow. 
We then only need to estimate our RGB camera's position and orientation, then reproject the 3D gait keypoints onto the images captured by those cameras as 2D pose.

Figure~\ref{fig:4_cam_view_indoor} shows an example set of 4 frames captured at the same time by 4 cameras. Figure~\ref{fig:indoor_accuracy} highlights the re-projection's pixel accuracy. We have a Vicon motion capture system for use. We will use "Vicon" and "motion capture system" to refer to our setup. 

\subsubsection{Preliminaries: Camera Model}
We use OpenCV's pinhole camera model. $\boldsymbol{K}$ represent the 3x3 camera intrinsic matrix. $\boldsymbol{d}$ represent the distortion coefficients. $\boldsymbol{R}\in SO(3)$ is the 3x3 rotation matrix representing camera's rotation in the world frame. $\boldsymbol{t}$ is the 3x1 vector that represent the camera's translation.  $\boldsymbol{R}$ and $\boldsymbol{t}$ together are called the extrinsic parameters of the camera. The 2D image point vector $\boldsymbol{p}_{2d}$ are 3x1 vector in  homogeneous coordinates and are assumed to have been normalized against its z value. 
\subsubsection{Camera Localization with PnP}
\label{subsubsec:pnp_localization}
We use the well-known perspective-n-point (PnP) computation to localize the cameras. The PnP method requires, as input, the above mentioned intrinsic parameters~$\boldsymbol{K}$,and distortion array $\boldsymbol{D}$, and a corresponding set of (3D,2D) pairs.

\noindent \textbf{Step 1 - obtain intrinsic parameters~$\boldsymbol{K}$, and distortion array $\boldsymbol{D}$} - We use Intel D455 cameras for  convenience as it provides its own intrinsic parameters. For an arbitrary pin-hole camera, one can use OpenCV's tool to estimate the their intrinsic parameters. It returns the camera's rotation $\boldsymbol{R}$ and its translation $\boldsymbol{t}$.

\noindent \textbf{Step 2 - obtain (3D,2D) pairs} -  Vicon already provides 3D data and we now need to manually mark the visible 2D markers on 1 frame with gait labels. Our semi-automated pipeline requires us to identify visible markers in an image manually which registers their image coordinates. This establishes several (3D,2D) pairs. In practice, we found that it is done best when the subject is facing directly towards/away from the camera with their arms extended. Then we can pick out around 10-20 pairs.

{\bf Note}: This manual step needs to be done only once, for each camera, for the entirety of the capture session. It would be ideal to tighten the cameras onto their tripod, record as many subjects as possible in one go. Any minor change in camera position requires the manual step to be repeated. In practice, we typically will perform this step only once for all data captured in a single day. 

\noindent \textbf{Step 3 - Camera Localization}
We now have ~$\boldsymbol{K}$,  $\boldsymbol{d}$ and the (3D,2D) pairs. We use PnP algorithm to obtain the camera's rotation $\boldsymbol{R}$ and its translation $\boldsymbol{t}$.

\subsubsection{3D-2D re-projection}
\label{subsubsec:3d_2d_reproject}
With camera intrinsics $\boldsymbol{K},\boldsymbol{d}$,and extrinsics $\boldsymbol{R},\boldsymbol{t}$, for any 3D point $\boldsymbol{p}_{3d}$, we can calculate its corresponding re-projected 2D coordinate $\boldsymbol{p}_{2d}$ in an image using perspective geometry: 

\begin{equation}
    \boldsymbol{p}_{2d}=\boldsymbol{K}\boldsymbol{R}^T(\boldsymbol{p}_{3d}-\boldsymbol{t})
\end{equation}
{\bf Note}: The obtained $\boldsymbol{p}_{2d}$ is in homogeneous coordinates and therefore must be normalized against its z value. 

We can now find all 3D point's 2D position on all images.

\subsubsection{Hardware setup, Timing and data synchronization}
Ideally, the motion capture system should be triggered at the exact same time as all the RGB cameras. In practice, this is quite challenging. The next best thing is to have each image timestamped by {\it ROS} and we manually observe 1 frame offset parameter between the RGB cameras and the motion capture system by going through the re-projected frame results. 

To achieve this, we use two computers synchronized with Network Time Protocol (NTP) and each is connected to two RGB cameras. We use a low-latency access point to establish a local area network that also connects to the Vicon work station. Using Vicon Nexus's software, one can trigger a UDP broadcast at the same moment as capture start/end. In practice, we observe that this still results in average 12-16 frames delay in the starting of RGB camera arrays. 

\subsection{Outdoor Setup and Annotation Pipeline}
In the outdoor settings, we do not have an off-the-shelf motion capture system. Instead, we first accurately estimate the position and orientation of 3 RGB close range cameras. We then use frames from 3 cameras with ~\href{https://developers.google.com/ml-kit/vision/pose-detection}{MLKIT} 2D pose label to triangulate each 3D gait keypoint. Lastly, we reproject the 3D gait keypoints onto the images captured by all cameras, including the 1 long range RGB camera as 2D pose. This 2D-3D-2D process is slightly different than before. 

{\bf Note}: It should be noted the MLKIT estimation of 2D pose on the long range camera is not possible since the subject appears as less than 25 pixels in height. See section~\ref{subsec:rgb_to_pose}.

\begin{figure}
    \centering
    \includegraphics[width=0.95\linewidth]{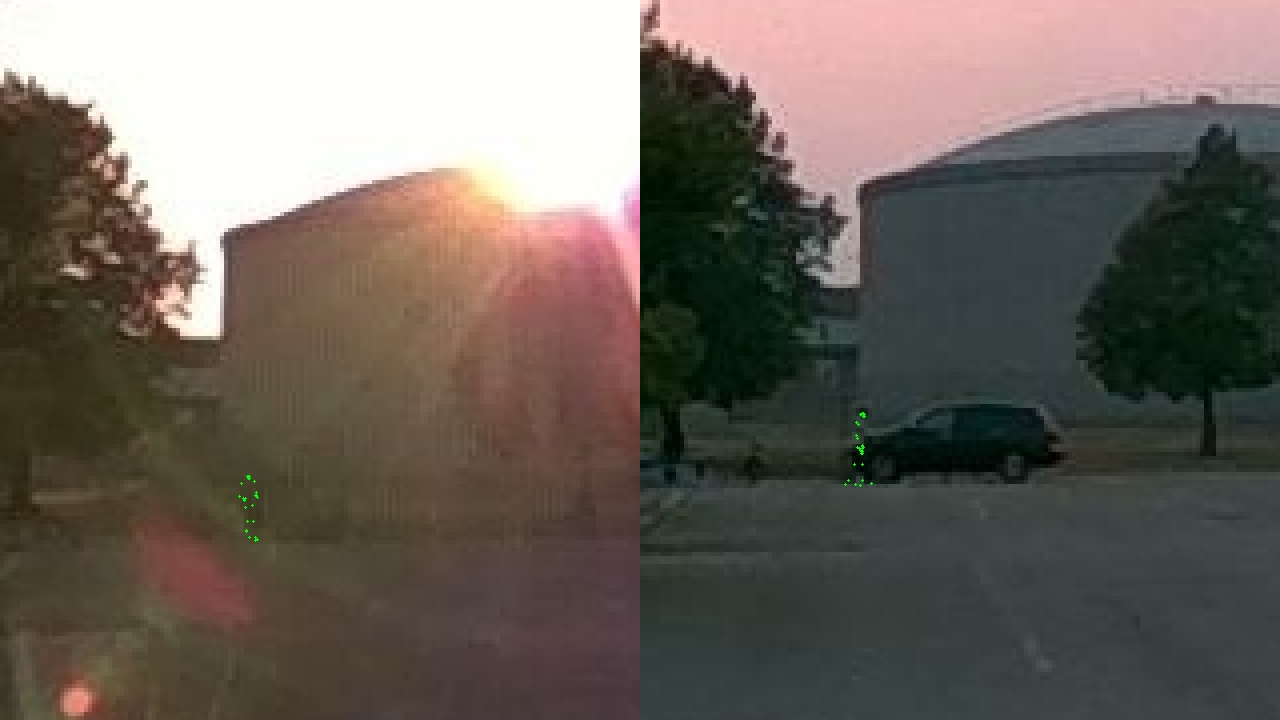}
\vspace{-7pt}    \caption{Long range captures under strong sun light exposure and partial occlusion. Note that the occluded scenario is for demonstration purpose, and not part of our data set.}
    
    \label{fig:occlusion_sun_exposure}
\end{figure}

{\bf Note}: For long range labeling, the method is robust to occlusion and direct sun exposure, since it only relies on the geometry relationship between the camera to reproject images. See Figure~\ref{fig:occlusion_sun_exposure}.

\subsubsection{Camera Array Localization with Bundle Adjustment}
\label{subsubsec:GTSAM}
To accurately localize cameras as well as ensure consistency across views, we utilize a well-known method called {\it bundle adjustment}~\cite{triggs2000bundle}~\cite{agarwal2010bundle}. It is the process of using image features across views to iteratively identify the camera location as well as corrected reprojection of the features for overall consistency. 
We use the GTSAM~\cite{gtsam}'s Bundle Adjustment library to localize our closed range camera array. This library requires, as input, each closed range camera's intrinsic parameters ~$\boldsymbol{K}_1,\boldsymbol{K}_2,\boldsymbol{K}_3...$, initial estimated orientation and position of each camera $(\boldsymbol{R}_1,\boldsymbol{t}_1),(\boldsymbol{R}_2,\boldsymbol{t}_2),(\boldsymbol{R}_3,\boldsymbol{t}_3)...$,  and a corresponding set of (3D,2D) pairs for each camera.  Unlike the above example, we use multiple frames here instead of only one.

The output is the optimized camera array extrinsic parameters - $(\boldsymbol{R}_1^*,\boldsymbol{t}_1^*),(\boldsymbol{R}_2^*,\boldsymbol{t}_2^*),(\boldsymbol{R}_3^*,\boldsymbol{t}_3^*)...$. 

\noindent \textbf{Step1: obtain camera intrinsic ~$\boldsymbol{K}$}
This is exactly the same as step1 of ~\ref{subsubsec:pnp_localization}

\begin{figure}
    \centering
    \includegraphics[width=0.9\linewidth]{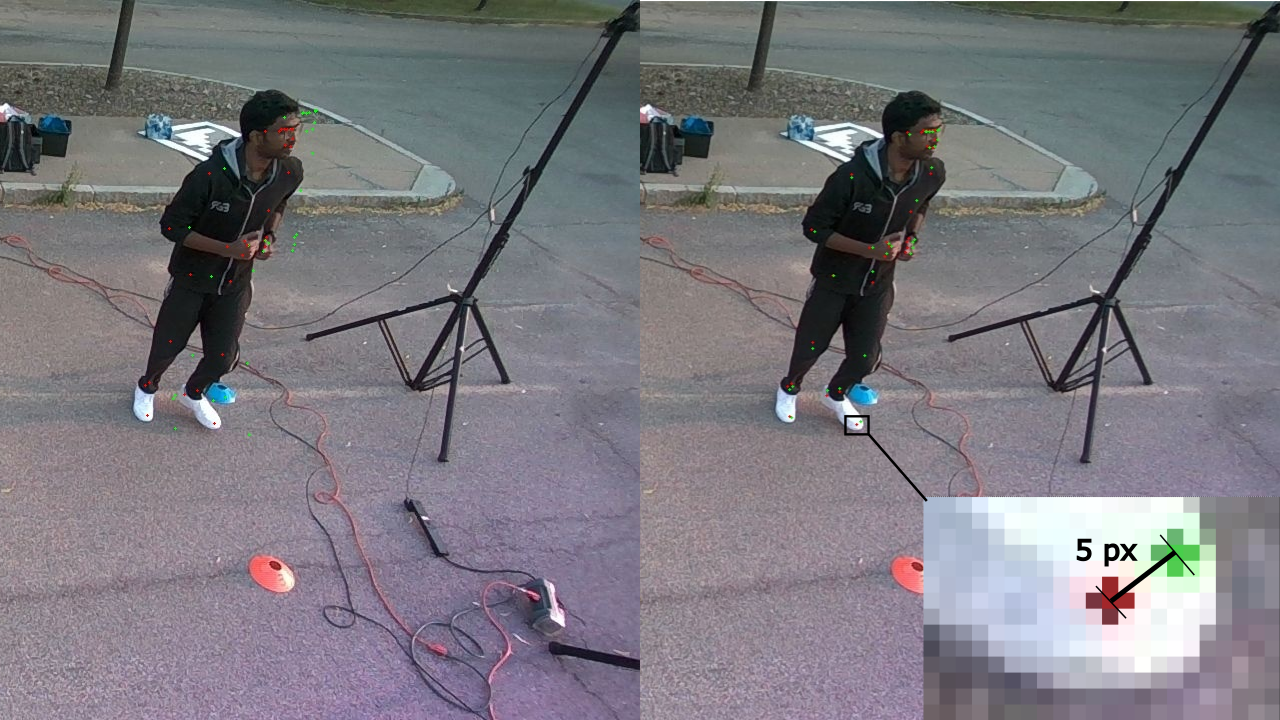}
    \caption{Highlighting the accuracy of camera extrinsic and 3D point position, after Bundle Adjustment. Zoom in to observe 2D gait keypoints. The red points are from MLKIT label. The green points are re-projected from the triangulated 3D pose. We observe an average of 5 pixel correction after bundle adjustment}
    \vspace{-10pt} 
    \label{fig:before_after_BA}
\end{figure}

\noindent \textbf{Step2: obtain initial estimations of camera extrinsics $(\boldsymbol{R}_1,\boldsymbol{t}_1),(\boldsymbol{R}_2,\boldsymbol{t}_2),(\boldsymbol{R}_3,\boldsymbol{t}_3)...$}
One can use a larger AprilTag for this task,
or, manual tape measurement. The point is the initial estimation does not have to be exactly correct, it will be optimized later by GTSAM. See figure~\ref{fig:before_after_BA}.

We use manual measurement of rough camera positions, with the assumptions that the cameras points at the center of the coordinate system. Through this, we have 3 measured camera translations $\boldsymbol{t}_1,\boldsymbol{t}_2,\boldsymbol{t}_3$, and we can obtain the orientation by:
\begin{gather}
    \label{eq:special-case-rotation}
\boldsymbol{R}_{i}
 =
  \begin{bmatrix}
   \cos{\theta_{a}} & -\sin{\theta_{a}}&0 \\
    \sin{\theta_{a}} & \cos{\theta_{a}} & 0 &\\
   0 & 0 & 1
   \end{bmatrix}
  \begin{bmatrix}
   1 & 0 & 0 &\\
    0 &\cos{\theta_{e}} & -\sin{\theta_{e}} \\
   0 & \sin{\theta_{e}} &\cos{\theta_{e}}   
   \end{bmatrix}
\end{gather}

where $\theta_{e}$ is the elevation angle:
\begin{equation}
    \theta_{e}=\frac{\pi}{2}-atan2(t_{iz},\sqrt{t_{ix}^2+t_{iy}^2})
\end{equation}

and $\theta_{a}$ is the azimuth angle:
\begin{equation}
    \theta_{a}=\frac{\pi}{2}-atan2(t_{iy},t_{ix})
\end{equation}

\noindent \textbf{Step 3: obtain 2D pose with MLKIT Learning Pipeline for short range cameras}: 
We first obtain the 2D pose for all images of each camera in a sequence, label each frame with MLKIT, then pick out the first 400 frames of each camera.  The frames from different cameras are synchronized by their time stamps. The 1200 total frames can triangulate 3D points in the first 400 time instances when the first 400 frames are captured by each camera. 

\color{black}

\noindent \textbf{Step 4: Obtain 3D pose by multi-views triangulation}: 
Various methods for triangulation exist, for clarity, we show our exact method here. Assume at a given time instance, a joint is marked by MLKIT in 3 different camera images as 3 2D points $\boldsymbol{p}_{2d1},\boldsymbol{p}_{2d2},\boldsymbol{p}_{2d3}$ where $1,2,3$ subscripts are the camera numbers. We also have each camera's intrinsic parameters ~$\boldsymbol{K}_1,\boldsymbol{K}_2,\boldsymbol{K}_3...$, initial estimated orientation and position of each camera $(\boldsymbol{R}_1,\boldsymbol{t}_1),(\boldsymbol{R}_2,\boldsymbol{t}_2),(\boldsymbol{R}_3,\boldsymbol{t}_3)...$, we need to find a single 3d point from these information $\boldsymbol{p}_{3d}$

Each 2d point $\boldsymbol{p}_{2di}$ marks a ray along with its camera's parameters, where the ray direction is 
\begin{equation}
    \boldsymbol{c}_i=\boldsymbol{R}_i\boldsymbol{K}_i^{-1}\boldsymbol{p}_{2di}
\end{equation}
And the ray origin is 

\begin{equation}
    \boldsymbol{k}_i=\boldsymbol{t}_i
\end{equation}

Thus with scalar $w_i$ the ray end point can be described as $\boldsymbol{k}_i+w_i\boldsymbol{c}_i$. Assume that the 3 rays intersect at one point:
\begin{equation}
    \boldsymbol{k}_1+w_1\boldsymbol{c}_1=\boldsymbol{k}_2+w_2\boldsymbol{c}_2=\boldsymbol{k}_3+w_3\boldsymbol{c}_3
\end{equation}

we can also write the above as three separate equations:

\begin{equation*}
    \boldsymbol{k}_1+w_1\boldsymbol{c}_1=\boldsymbol{k}_2+w_2\boldsymbol{c}_2
\end{equation*}
\begin{equation*}
    \boldsymbol{k}_1+w_1\boldsymbol{c}_1=\boldsymbol{k}_3+w_3\boldsymbol{c}_3
\end{equation*}
\begin{equation*}
    \boldsymbol{k}_2+w_2\boldsymbol{c}_2=\boldsymbol{k}_3+w_3\boldsymbol{c}_3
\end{equation*}

We convert the above 3 equations system into its $\boldsymbol{A}\boldsymbol{x}=\boldsymbol{b}$ matrix form:
\begin{gather}
    \label{eq:special-case-rotation}
  \begin{bmatrix}
   \boldsymbol{c}_1 & -\boldsymbol{c}_2 &0\\
    \boldsymbol{c}_1 &0& -\boldsymbol{c}_3 \\
   0 & \boldsymbol{c}_2 & -\boldsymbol{c}_3
   \end{bmatrix}
  \begin{bmatrix}
   w_1 \\
    w_2  \\
   w_3    
   \end{bmatrix}
   =
  \begin{bmatrix}
   \boldsymbol{k}_2-\boldsymbol{k}_1 \\
    \boldsymbol{k}_3-\boldsymbol{k}_1  \\
   \boldsymbol{k}_3-\boldsymbol{k}_2    
   \end{bmatrix}
\end{gather}

Note that $\boldsymbol{c}_i,\boldsymbol{k}_i$ are 3x1 column vectors, the above matrix has dimensions (9x3)(3x1)=(9x1).

We can now use linear least squares to find $\boldsymbol{x}=[w_1,w_2,w_3]^T$:
\begin{equation}
    \boldsymbol{x}=(\boldsymbol{A}^T\boldsymbol{A})^{-1}\boldsymbol{A}^T\boldsymbol{b}
\end{equation}

with $[w_1,w_2,w_3]^T$ found, we can then take the average of 3 end points as the triangulated $\boldsymbol{p}_{3d}$:
\begin{equation}
    \boldsymbol{p}_{3d}=\frac{1}{3}(    \boldsymbol{k}_1+w_1\boldsymbol{c}_1+\boldsymbol{k}_2+w_2\boldsymbol{c}_2+\boldsymbol{k}_3+w_3\boldsymbol{c}_3)
\end{equation}

\noindent \textbf{Step 5: Refined camera extrinsics with GTSAM Bundle Adjustment}: We now have the required inputs, namely ~$\boldsymbol{K}_1,\boldsymbol{K}_2,\boldsymbol{K}_3...$, initial estimated orientation and position of each camera $(\boldsymbol{R}_1,\boldsymbol{t}_1),(\boldsymbol{R}_2,\boldsymbol{t}_2),(\boldsymbol{R}_3,\boldsymbol{t}_3)...$,  and a corresponding set of (3D,2D) pairs for each camera.  We can now use GTSAM(\href{https://gtsam.org/tutorials/intro.html}{SfM}) to obtain the optimized camera array extrinsic $(\boldsymbol{R}_1^*,\boldsymbol{t}_1^*),(\boldsymbol{R}_2^*,\boldsymbol{t}_2^*),(\boldsymbol{R}_3^*,\boldsymbol{t}_3^*)...$

%\todo[inline]{"Check GTSAM's tutorial, SfM example for this step." - was this for your reference? You don't need to say this in the paper. }
For a comparison of before/after bundle adjustment, see Figure~\ref{fig:before_after_BA}.

% \todo[inline]{check if 5 px is the right number in figure~\ref{fig:before_after_BA} You can give an approx number - don't need to calculate actual one. Chen: Yes, 5 px is exactly correct, the other 2 sides of the triangle are counted as 4px and 3px }

\noindent \textbf{Step 6: Triangulate 1 more time for optimized 3d points}: Repeat step 4 but with optimized camera extrinsics.

\noindent \textbf{Step7: 3D-2D re-projection for all cameras}: 
This step is exactly the same as Section~\ref{subsubsec:3d_2d_reproject}, but uses the optimized camera extrinsics here and optimized 3d points
\subsubsection{Long Range Camera Initial extrinsic and manual refinement}
\begin{figure}
    \centering
    \includegraphics[width=0.9\linewidth]{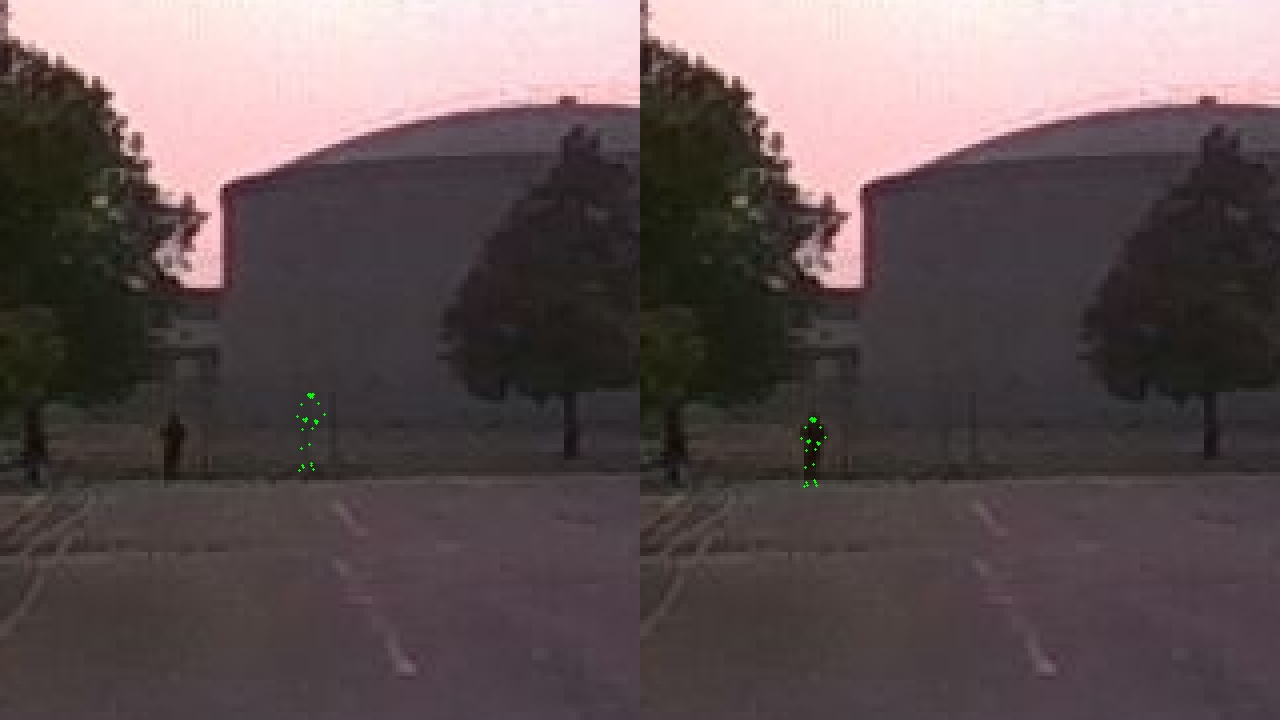}
    \caption{Before(left) and after(right) long range camera extrinsic manual adjustment.}
    \vspace{-10pt} \label{fig:long_range_cam_adjustment}
\end{figure}

After we have optimized the close ranged camera's extrinsics, we can start refining the long range camera's extrinsic. Closely related to step 2 in Section~\ref{subsubsec:GTSAM}, we can add/subtract $\Delta\theta$ from $\theta_{e}$ or $\theta_{a}$, which has the effect of adjusting the reprojected skeleton in the vertical and horizontal images coordinates, respectively, until the reprojected skeleton aligns. It is recommended during the setup process to place the long range camera on one of the coordinate's plane axis, X or Y. In our setup we chose the positive Y-axis. With such setup,  we can also add/subtract a small $\Delta d$ in meters, to the long distance axis of $\boldsymbol{t}$. changing this variable has the effects of shrinking/magnifying the skeleton on image. For an example of such adjustment, see Figure~\ref{fig:long_range_cam_adjustment}.

 % Empiracally we observed this step is enough for accuractely reprojecting the 3d pose to long range 2d image. We will show this during evaluation and with videos.
{\bf Note}: This procedure only has to be done once per collection day.

%% file: eval.tex
\section{Evaluation}
\label{sec:eval}
In this section, we evaluate our 3D/2D pose quality in three seperate parts -- the quality of indoor 2d pose data; outdoor closed range 2d/3D pose data; and outdoor long range 2D/3d pose data. 

For each part we present a quantitative metric. In multi-view camera  evaluation,  the accuracy of the 3D/2D points is typically presented as 3 re-projected 2D pixel residual error in pixels, in short, reprojection error.

\begin{table*}[h!!]
  \centering
  \begin{tabular}{|ccccccccccc|}
    \hline
    Pipeline & Pretraining & Evaluation & In-1 & In-2 & In-3 & In-4 & Out-1 & \textbf{Out-2 (long range)} & Out-3 & Out-4 \\
\specialrule{1.5pt}{0pt}{0pt} 
    GaitGraph& CASIA-B& DIOR&8.418	&8.418	&8.418	&10.20	&6.888	&8.163	&13.52	&11.22\\
    GaitMixer& CASIA-B& DIOR&5.612	&7.653	&6.378	&7.908	&8.418	&8.418	&10.20	&12.50\\  
    GaitFormer& CASIA-B& DIOR&6.888	&7.653&5.867	&6.633	&9.949	&7.143	&11.99&10.71\\
    \hline
  \end{tabular}
  \caption{Cross-domain Rank@1 results from 3 pipelines. The network weights are obtained from author provided checkpoints without additional training. "in-n" are indoor camera numbers. "out-n" are outdoor camera numbers. "out-2" is the long range camera, it's accuracy is reflective of using the long range camera sequences as probe, and the rest as gallery.}
  \vspace{-7pt}
  \label{tab:reid}
\end{table*}

\subsection{Indoor Data Evaluation}

%\color{blue}

% on the accuracy on the triangulated 3D points and the camera extrinsics are typically based on "re-projection error" which is the pixel difference between observed 2d key point and the reprojected 2d keypoint.   

\subsubsection{Reprojection Error}

\begin{figure}
    \centering
    \includegraphics[width=0.5\linewidth]{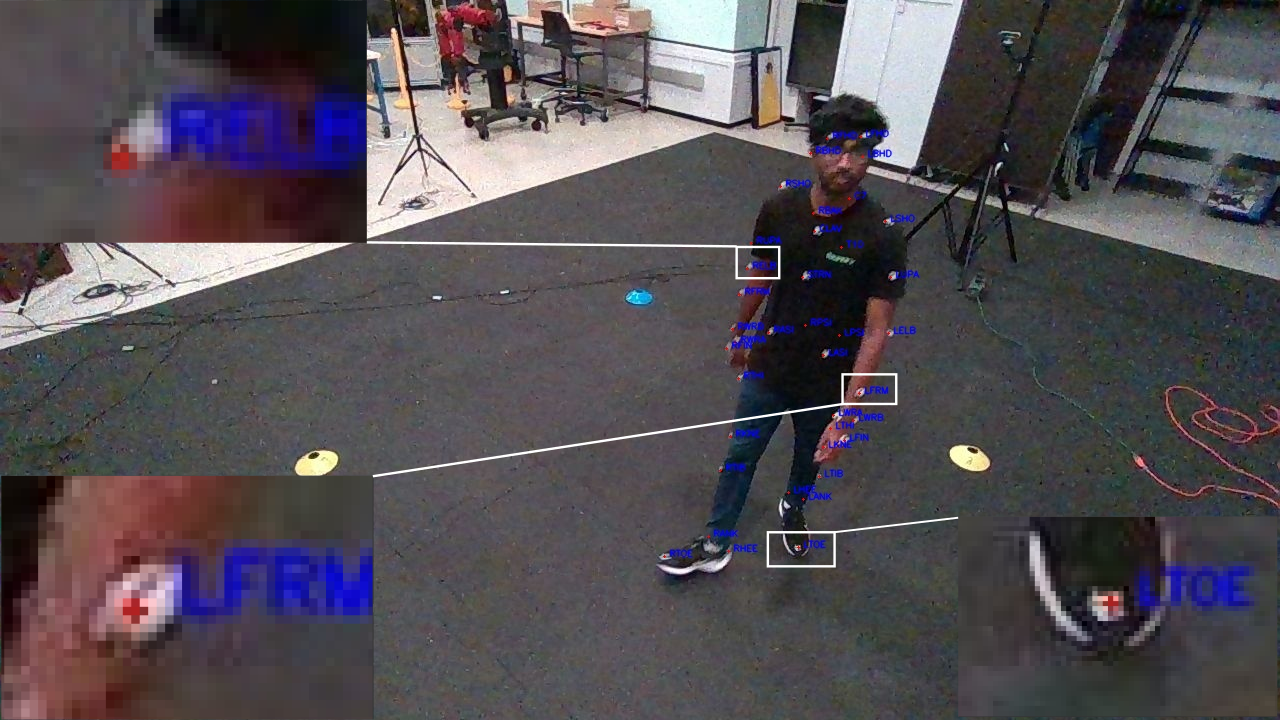}
    \caption{This picture highlights the label's pixel accuracy in the indoor environment. The reflective markers are white in color, the 2d labels are in red dots, their names are in blue text. }
    \vspace{-10pt}
    \label{fig:indoor_accuracy}
\end{figure}

\begin{figure}
    \centering
    \includegraphics[width=0.95\linewidth]{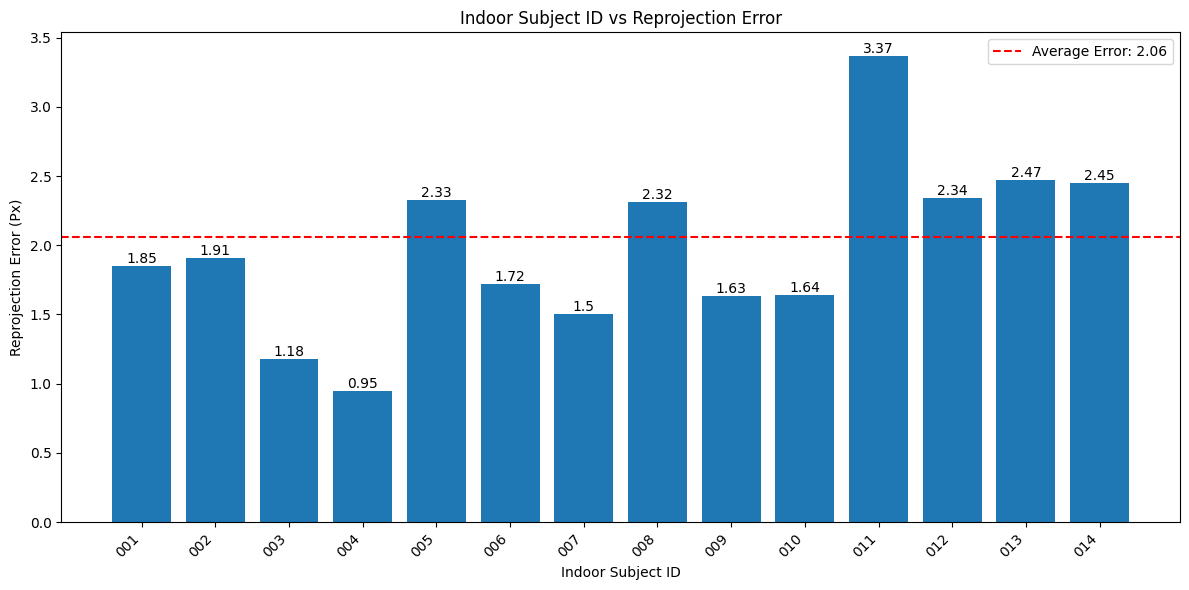}
    \caption{Reprojection error in the indoor motion capture environment.To get a sense of image distance by pixel, see a 5 pixel residue example in Section~\ref{fig:before_after_BA}. }
    \label{fig:indoor_reprojection_error}
    \vspace{-7pt}
\end{figure}

The 3D data captured by a motion capture system are accurate to the mm level and are typically regarded as ground truth. Therefore, we only  evaluate the 2d pose accuracy. 
To do so, we first randomly select 100 2D points from the data set to manually label, then compare to the corresponding 2d labels from 3D-2D projection ~\ref{subsubsec:3d_2d_reproject}. Then we use the same reprojection error Equation~\ref{eqn:reprojectionerror}. The re-projection error here is the difference between the re-projected 2D pose from vicon $\boldsymbol{p}_{vicon}$ and randomly selected manual
label 2D pose $\boldsymbol{p}_{2d}$. The result is shown in Figure~\ref{fig:indoor_reprojection_error}.

\subsection{Outdoor Evaluation}
\subsubsection{Infeasibility of Long-Range Detection}
For outdoor data, a mechanism to identify 2D gait keypoints is to apply a standard gait recognition algorithm on the long range camera view.  We evaluated 3 different pipelines as is, without additional training. They are AlphaPose, MLKIT, YOLOv7. We evaluate on both the original long range RGB images, and a cropped, zoomed-in version of the RGB image. The evaluation is done to the entire set of 211798 long range raw images and the zoomed in version of the images.  The result in shown in Table~\ref{tab:2} and Table~\ref{tab:3}. From Tables~\ref{tab:2},~\ref{tab:3}, we can see that the detection accuracy is 12\% for MLKIT and goes down to 0\% for correct detections. Therefore, we determined that labeling long range views directly using a gait pipeline was not feasible. We opted to use close range cameras and reprojection from the close cameras to the long range camera for gait labeling in the long range view. 
\label{sec:baseline}
\begin{table}
  \centering
  \begin{tabular}{|ccc|}
    \hline
    Pipeline &raw detection  &zoomed detection \\
\specialrule{1.5pt}{0pt}{0pt} 
    MLKIT& 12.32& 1.02\\
    YOLOv7& 1.15& 2.03\\
    AlphaPose& 6.84& 4.60\\
    \hline
  \end{tabular}
  \caption{Across the entire set of 211798 long range frames, The first column is the percentage of detection over none zoom-in images. The second column is the percentage of detection over zoomed in images. An example of zoomed-in frame is shown in figure~\ref{fig:long_range_IOU}. Which is a 720p frame zoomed in 4X towards the image center, with raw dimension 320x180}
  \label{tab:2}
\end{table}
\begin{table}
  \centering
  \begin{tabular}{|ccc|}
    \hline
    Pipeline &raw success &zoomed success \\
\specialrule{1.5pt}{0pt}{0pt} 
    MLKIT& 0.0& 0.0\\
    YOLOv7& 0.0& 0.005\\
    AlphaPose& 5.57& 0.36\\
    \hline
  \end{tabular}
  \caption{Different than table~\ref{tab:2}, we evaluate the detected points against our skeleton gait keypoint's bounding box. If 20 percent of the pipeline detected keypoints fall within this bounding box then we count the detection as succesful. This criteria is rather generous.}
  \label{tab:3}
  \vspace{-15pt}
\end{table}

\label{subsec:rgb_to_pose}

%We will separate the evaluations of the closed range cameras and the long range camera as there are no MLKIT labels on the long range camera. 

\label{sec:results}
% Make sure that the Paper ID from the submission system is visible in the version submitted for review (replacing the ``*****'' you see in this document).
% If you are using the \LaTeX\ template, \textbf{make sure to update paper ID in the appropriate place in the tex file}.

\subsubsection{Close Range Cameras Reprojection Error}
\begin{figure}[h]
    \centering
    \includegraphics[width=0.9\linewidth]{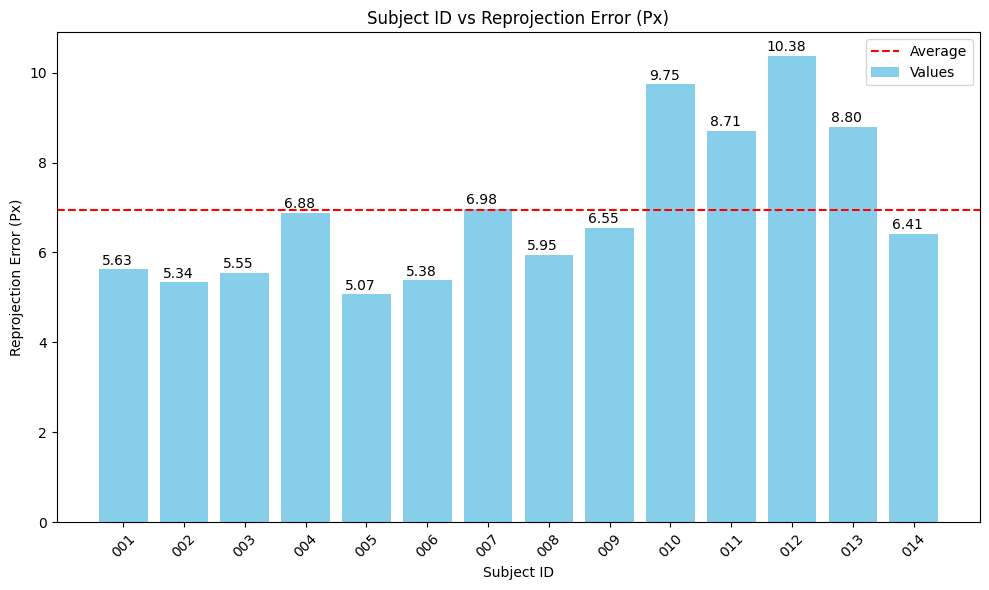}
    \caption{Evaluation of reprojection error, over the entire set of 600k+ out outdoor closed range camera images. To get a sense of image distance by pixel, see a 5 pixel residue example in ~\ref{fig:before_after_BA}.}
    \vspace{-10pt}
    \label{fig:reprojection_error_by_id}
\end{figure}

A challenge in multi-camera gait keypoint recognition is the error across the multiple views. We will use the re-projection error to evaluate the accuracy of outdoor closed range camera extrinsics, as well as the triangulated 3D points. The re-projection error here is the difference between the re-projected 2D pose $\boldsymbol{p}_r$ and the MLKIT label 2D pose $\boldsymbol{p}_{2d}$.  The result is plotted in Figure~\ref{fig:reprojection_error_by_id}.

\begin{equation}
    e=\frac{1}{n}\sum_{n}\sqrt{(\boldsymbol{p}_r-\boldsymbol{p}_{2d})^T(\boldsymbol{p}_r-\boldsymbol{p}_{2d})}
    \label{eqn:reprojectionerror}
\end{equation}

\subsubsection{Far Range Camera Accuracy}
\begin{figure}
    \centering
    \includegraphics[width=0.9\linewidth]{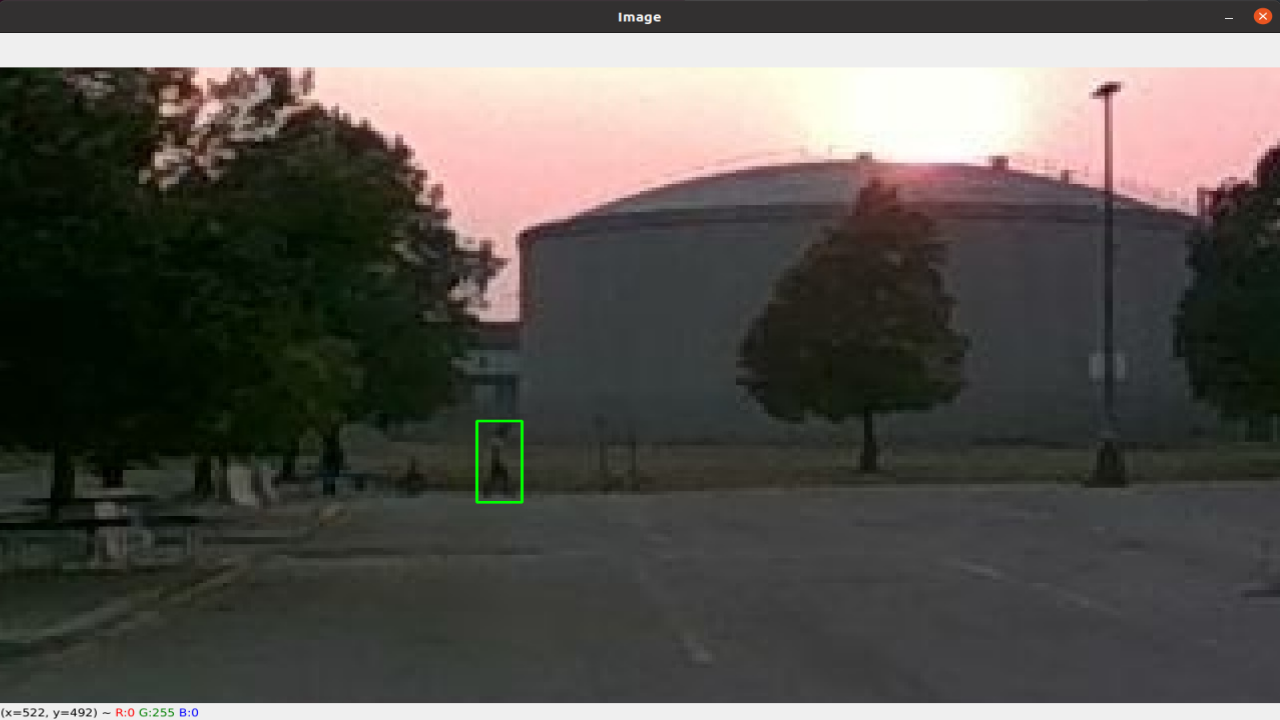}
    \caption{We manually label 14245 frames with bounding boxes for our quantitative evaluations. The picture shows the labeling interface.}
    \label{fig:long_range_IOU}
    \vspace{-11pt}
\end{figure}
\begin{figure}
    \centering
    \includegraphics[width=0.95\linewidth]{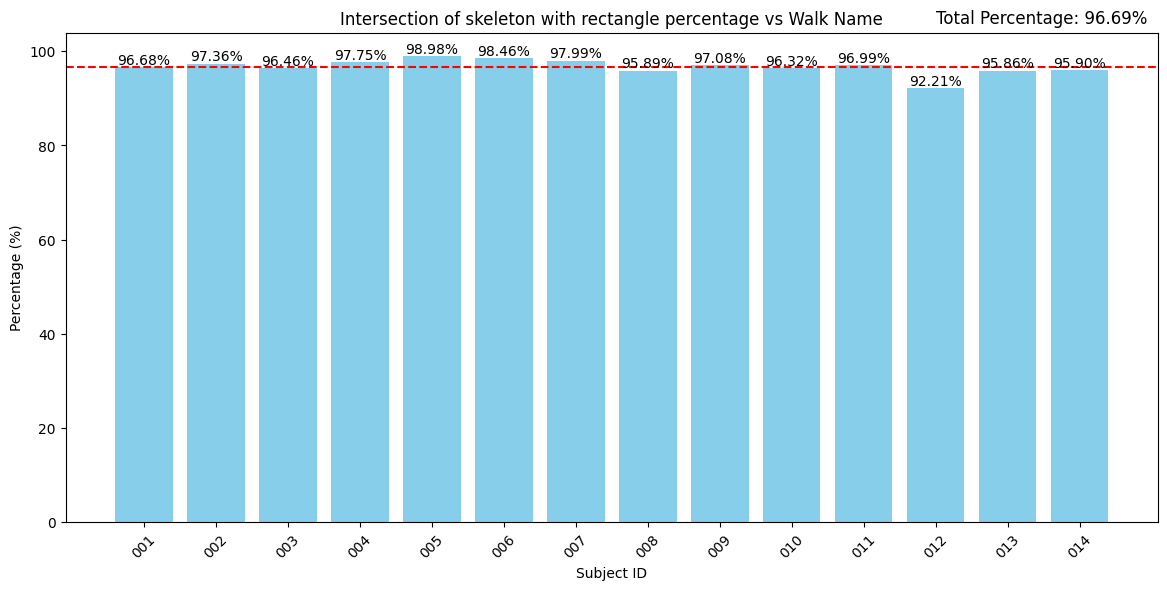}
    \vspace{-7pt}
    \caption{Across the entire set of randomlly selected 14245 long range images, 96.69 percent of the long range 2d pose falls within the manual bounding boxes.}
    \vspace{-10pt}
    \label{fig:iou_by_id}
\end{figure}

We use a coarser form of re-projection error on the long range camera as there are no MLKIT labels. Manual labeling of each join point is difficult and ineffective on subjects that appear in such low pixel counts. Instead, 
we will manually select  rectangles 14245 random far range frames, and count the percentage of 2D labels that falls into the bounding box of the subject.
See Figure~\ref{fig:long_range_IOU} for an example of using a script to selecting a rectangle on 1 frame. Figure~\ref{fig:iou_by_id} shows percentage of gait keypoints within label rectangles, split by subject id.

\subsubsection{Gait Recognition (Re-ID)}

We evaluate GaitGraph~\cite{teepe2021gaitgraph},GaitMixer~\cite{pinyoanuntapong2023gaitmixer} and GaitFormer~\cite{pinyoanuntapong2023gaitmixer}, with checkpoints trained on casia-b dataset. In this cross-domain settings, we use sequences captured from indoor setttings and closed range camera in outdoor settings as the gallery, and using long range camera sequences as probe. we report Rank-1 subject re-identification accuracy. The result can be seen in Table~\ref{tab:reid}. All ReID evaluations are performed using 300 frame sequences. As it can observed in Table~\ref{tab:reid}, for long-range scenario (out-2), GaitMixer provides the best performance, followed by GaitGraph and GaitFormer.

%% file: dataset.tex
\section{Dataset}
\label{sec:dataset}
\subsection{Statistics and collection protocol}
There are 1,649,918 total frames in our dataset, where 802726 frames are from indoor MoCap settings, and 847193 frames from ourdoor settings. There are a total of 59,530,485 2d gait key-points. And approximately 14.88 millions of 3d gait keypoint.
For the indoor settings, we additionally have each subject's 360degree profile, from two camera angles, frontal and 45 degrees downwards, for a total of around 20,000 images. 
For the outdoor setting's 847193 frames, 211798 frames are from the long range camera.

There are 14 subjects and 112 sequences total. 56 indoor sequences from 14 subjects with 2 walk patterns and 2 outfits. 56 outdoor sequences from the same 14 subjects with 2 walk patterns and 2 outfits. 

Each walk lasts 2 minutes, with 30FPS capture rate, thus yields around 3600 frames on each camera. There are 4 cameras in both indoor and outdoor settings. In the outdoor setting 1 camera is place at approximately 60 meters. on 720p frame, subject apears as 20-25 pixels in height.

\subsection{Annotation Time Cost}
The indoor 800k frames took approximately 77hrs.The outdoor 800k frames, including the 200k long range frames, took approximately 42 hrs.The aggregated per frame time cost is therefore around .2678 second.

%% file: conc.tex
\section{Conclusion}
%We would like to acknowledge limitations of our study:
%\begin{itemize}
%    \item Requirement of short range camera arrays in the outdoor long range settings, which can limit the use of this approach.
%\end{itemize}

%In this work we outline our approach to collect and auto-annotate 3D/2D pose in the indoor and outdoor settings, with long range cameras.
This work presents DIOR, a first of its kind dataset with indoor and outdoor data of 14 subjects performing various walking and running activities in a limited space. The subjects are also seen with multiple pieces of clothing. Each image is annotated with gait keypoints for use by algorithms that can use this anthropomorphic information. The dataset and pipeline will be published openly for others to use upon publication.